\documentclass[10pt,twocolumn,letterpaper]{article}

\usepackage{iccv}
\usepackage{times}
\usepackage{epsfig}
\usepackage{graphicx}
\usepackage{amsmath}
\usepackage{amssymb}
\usepackage{algorithm}
\usepackage{algpseudocode}
\usepackage{multirow}

\usepackage[breaklinks=true,bookmarks=false]{hyperref}

\iccvfinalcopy 

\begin{document}

\title{SRIL: Selective Regularization for Class-Incremental Learning}

\author{Jisu Han \qquad Jaemin Na \qquad Wonjun Hwang\\
Ajou University\\
\{jisu3709, osial46, wjhwang\}@ajou.ac.kr\\}

\maketitle

\begin{abstract}
Human intelligence gradually accepts new information and accumulates knowledge throughout the lifespan. However, deep learning models suffer from a catastrophic forgetting phenomenon, where they forget previous knowledge when acquiring new information. Class-Incremental Learning aims to create an integrated model that balances plasticity and stability to overcome this challenge. In this paper, we propose a selective regularization method that accepts new knowledge while maintaining previous knowledge. We first introduce an asymmetric feature distillation method for old and new classes inspired by cognitive science, using the gradient of classification and knowledge distillation losses to determine whether to perform pattern completion or pattern separation. We also propose a method to selectively interpolate the weight of the previous model for a balance between stability and plasticity, and we adjust whether to transfer through model confidence to ensure the performance of the previous class and enable exploratory learning. 
We validate the effectiveness of the proposed method, which surpasses the performance of existing methods through extensive experimental protocols using CIFAR-100, ImageNet-Subset, and ImageNet-Full. \footnote{Our code will be available at https://github.com/anonymous.}
\end{abstract}

\section{Introduction}

The successful development of deep learning has created many businesses and has applied to the real-world. However, many studies on deep learning have been evaluated in limited experimental settings. To achieve human-level goals, consideration of changes in real-world environments is required. In this respect, continual learning has recently been receiving a lot of attention from the artificial intelligence community. Continual learning is a methodology that can continuously learn about real-world situations where visual characteristics change, such as robot vision and autonomous driving systems~\cite{autonomous}. Although recent studies on continual learning have shown promising results, they still suffer from the problem of losing previous knowledge in the process of learning new knowledge. This phenomenon is called catastrophic forgetting~\cite{Goodfellow13,mccloskey1989catastrophic}, and this problem results in performance degradation of deep learning models. One of the major concerns of recent continual learning approaches is mitigating the catastrophic forgetting problem to prevent such performance deterioration.

\begin{figure}
\begin{center}
\includegraphics[width=8cm]{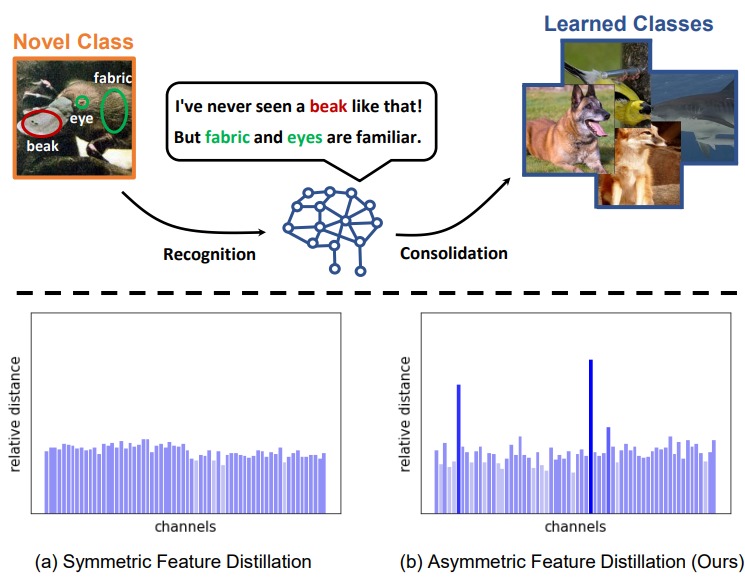}
\end{center}
   \caption{The illustration above shows our simple intuition of class-incremental learning. Our intuition is that in the process of learning a new class, there are new representations (e.g., beak) to learn to improve the overall classification performance, and there are representations that need to be differentiated while preserving  previous knowledge (e.g., fabric and eyes). The histogram below shows the change for each channel of the intermediate feature when a model that has pretrained 50 classes learns 1 additional class. Unlike previous feature distillation methods, we have a few dominant channels for pattern separation.}
\label{fig:Figure1}
\vspace{-0.4cm}
\end{figure}

Class-Incremental Learning (CIL) is a special case of the continual learning that seeks to learn new classes that have not been learned before while preserving knowledge from old classes. To this end, recent works have grafted knowledge distillation techniques onto CIL and have achieved considerable results. The major approaches for applying knowledge distillation to CIL can be divided into response-based methods~\cite{ICaRL,EEIL,BiC}, relation-based methods~\cite{TPCIL,ERDIL}, and feature-based methods~\cite{LUCIR,PODNet,AFC}. Both the response-based and relation-based methods transfer the dark knowledge of the old model considering different characteristics of the semantic information through asymmetric learning strategies for old and new data~\cite{ERACE, OCM}. On the other hand, the feature-based method directly transfers representation of the old model, making it difficult to consider the different characteristics of the old and new classes.
In this paper, we introduce a selective regularization method named \textbf{SRIL} that leverages knowledge distillation and weight interpolation.
First, we propose an asymmetric feature distillation method called \textbf{Gradient-based Feature Distillation}. Our intuition begins with pattern completion and pattern separation in cognitive science~\cite{patternseparation1, patternseparation2}. The pattern completion means integrating memories by judging that they are existing knowledge when they are newly readjusted by external stimuli, and pattern separation is the process of making memories distinct from existing memories by judging that they are different from the contents in memory. We intend to achieve realignment of knowledge by adopting an asymmetric learning strategy for the previously learned class and the newly learned class in terms of pattern completion and pattern separation. In the process of learning a new class, we utilize the gradient of the knowledge distillation and classification losses to determine whether knowledge distillation is beneficial or harmful. Then, we generate a mask for the channel of each intermediate feature and exploit the mask to apply feature distillation to conflicting channels for the old and new classes. Selective feature distillation through masks generated by this process considers different characteristics by taking an asymmetric strategy for old and new classes. As a result, Figure~\ref{fig:Figure1} shows the difference between symmetric feature distillation and our proposed asymmetric feature distillation. Our method has large feature changes for few dominant channels for pattern separation and small feature changes for pattern completion for the remaining channels.
Meanwhile, the stability-plasticity dilemma~\cite{mermillod2013stability,verbeke2019learning} of forgetting old knowledge while learning new knowledge, and failing to accept new knowledge in order to not to forget the old knowledge, is a well-known problem in CIL. To deal with this issue, we introduce a \textbf{Confidence-aware Weight Interpolation}, which determines whether the model retains old knowledge well and performs selective regularization according to the determined result. Here, we determine whether the new model maintains previous knowledge through the confidence of the new model on old data. The existing weight interpolation method is used to improve the generalization performance of a model as a method for approximating an ensemble~\cite{WiSEFT,modelsoups}. In contrast, we ensure that the upper bound of the loss for the old data does not increase by making the new model close to the old model in the weight space to ensure model stability. However, not to move away from the weight space can rather be a constraint on learning new knowledge. Therefore, if the new model's confidence in the old data exceeds a certain value based on the old model's confidence, we remove regularization to enable exploratory learning and balance stability and plasticity.
Overall, our contributions are as follows:
\begin{itemize}
    \setlength\itemsep{-0.1em}
    \item We propose a new asymmetric Gradient-based Feature Distillation (GFD) to consider feature characteristics between old and new classes. 
    \item We propose Confidence-aware Weight Interpolation (CWI) to strike a balance between stability and plasticity. CWI ensures the stability of the model and enables exploratory learning by operating selectively.
    \item We achieve comparable performance to the recent state-of-the-art methods and validate our methods through various experimental settings in CIFAR-100, ImageNet-Subset, and ImageNet-Full.
\end{itemize}
\begin{figure*}
\begin{center}
\includegraphics[width=13cm]{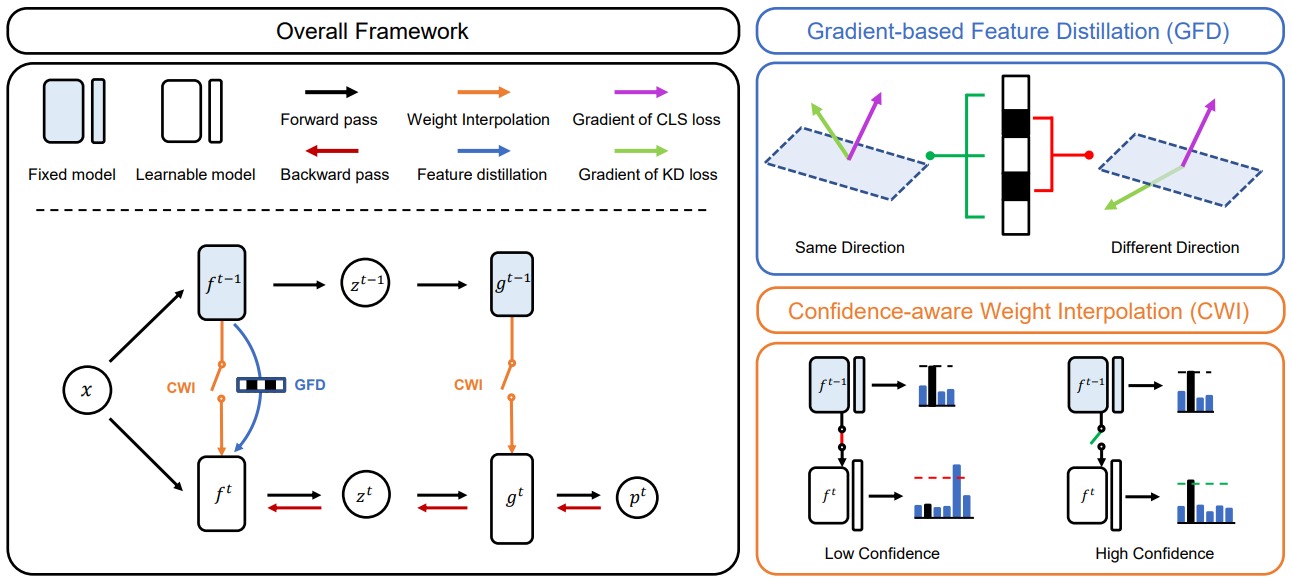}

\vspace{-4mm}
\end{center}
   \caption{Overview of the selective regularization framework SRIL. Our framework consists of two methods: GFD and CWI. \lowercase\expandafter{(\romannumeral1)} GFD calculates the cosine similarity between gradients of classification loss and knowledge distillation loss to generate a binary mask that applies knowledge distillation when in the same direction and does not apply when in different directions. \lowercase\expandafter{(\romannumeral2)} CWI determines whether to apply weight interpolation based on the confidence of the current model for old data.}
\label{fig:Figure2}
\vspace{-0.4cm}
\end{figure*}

\section{Related Work}
\textbf{Class-Incremental Learning.} 
CIL is largely classified into three types according to how to solve the problem. \lowercase\expandafter{(\romannumeral1)} Regularization based methods regularize the difference between the parameters of the previous network and current network to maintain a low loss area for previous tasks~\cite{EWC,RWALK,MAS}. \lowercase\expandafter{(\romannumeral2)} Dynamic architecture method is a method that continuously adds sub-networks or expands the network dynamically through pruning and retraining~\cite{PNN,DEN,CPG}. \lowercase\expandafter{(\romannumeral3)} Replay based method allows a limited amount of memory composed of data from previous tasks. As a representative method, studies are being conducted to transfer dark knowledge of previously trained networks using knowledge distillation~\cite{ICaRL,EEIL,LUCIR} or to improve performance through efficient memory management strategies~\cite{Mnemonics,RMM}.

\textbf{Knowledge Distillation.}
Knowledge distillation (KD) approaches have been proposed to improve the performance of lightweight models by transferring knowledge from large models to small models~\cite{KD}. Similarly, many studies have been conducted in CIL to solve the catastrophic forgetting problem by transferring the knowledge of the previous model through KD when learning about new data. LwF~\cite{LwF} was the first paper to apply KD to CIL, and uses KD so that the output of the model follows the output of the previous trained model. iCaRL~\cite{ICaRL} used exemplar to store a small amount of old data samples and applies KD to new data to maintain representation. LUCIR~\cite{LUCIR} proposed feature distillation in the embedding space rather than prediction by logit. PODNet~\cite{PODNet} applied distillation not only in the embedding space but also in the intermediate features. GeoDL~\cite{GeoDL} also considered the learning path on sub dimension in the KD process through an additional regularization term for the geodesic between features in the embedding space. AFC~\cite{AFC} defined the gradient magnitude of each channel in intermediate features as importance and applies differentiated feature distillation for each channel.

\textbf{Weight Space Ensemble.}
Ensemble is the most basic method used to improve the performance of a model in deep learning. However, existing ensemble methods are accompanied by an increase in model size or additional cost in the inference process. SWA~\cite{SWA} showed that by performing weight averaging periodically along the trajectory of SGD in a single model, the model improves in more generalized performance and converges to a wide minima. WiSE-FT~\cite{WiSEFT} demonstrated that the weight space ensemble from the zero-shot model improves the robustness of the model in the fine-tuning process. In a recent study of model soups~\cite{modelsoups}, the performance of the model was improved without additional inference time by applying the weight space-ensemble from the fine-tuned multiple model. Inspired by these weight space ensemble methods, there have been attempts to apply them in CIL. It either transfer the knowledge of new classes to the old model by updating the old model via exponential moving average~\cite{stojanovski2022momentum}, or apply consistent weight interpolation for the entire learning process.~\cite{eeckt2022weight}.

\section{Method}
\subsection{Problem Definition}
CIL aims to improve performance for all classes by starting from learning a limited number of classes and sequentially adding new classes. These sequential stages are called tasks. In CIL scenario, we have sequential training dataset $D^{t}=\left\{ x^{t}_i,y^{t}_i \right\}^N_{i=1}$ and exemplar $\mathcal{E}^{t}=\left\{ x^{1:t-1}_j,y^{1:t-1}_j\right\}^R_{j=1}$ containing limited data for previous tasks, where $t$, $x$ and $y$ denote task, input image and labels, respectively, and $1:t$ denotes from the $1$-st task to the $t$-th task. Each task's classes are disjoint in CIL scenario, $i.e., \ y^{1:t-1} \cap y^{t} = \varnothing$. Since our goal is to maintain the performance of previous tasks while learning new tasks, the optimization problem is defined as follow:
\begin{align}
\begin{split}
\arg\min_{\theta} \ \mathbb{E}_{(x,y)\sim D^{1:t}} \left[\mathcal{L}\left( x,y;\theta \right)\right]\\
s.t. \  \mathbb{E}_{(x,y)\sim D^{t} \cup \mathcal{E}^{t}} \left[\mathcal{L}\left( x,y;\theta \right)\right] < \epsilon,
\end{split}
\end{align}
where the $\mathcal{L}$ is the loss function (e.g., Cross-Entropy), and the $\theta$ is parameters of the model. Since the model converges while learning from current task data and exemplar, there exists some small number $\epsilon$ higher than loss.
\subsection{Gradient-based Feature Distillation}
We use the gradients of classification and KD losses for pattern integration and separation, inspired by ~\cite{cos,SCKD}. We consider that each channel contains information about patterns in the features of the model, therefore we decide whether to apply feature distillation to each channel through the similarity between the gradient of the classification and the KD losses. To achieve this, we build a binary mask $M$ whose activation is determined by the cosine similarity of the two gradients.
\begin{align}
M_{l,c} = \begin{cases}
1,\; if\; \; cos(\nabla_{z_{l,c}} \mathcal{L}_{kd} , \nabla_{z_{l,c}} \mathcal{L}_{cls})\ge 0 \\
0,\; \mathrm{otherwise}
\end{cases},
\end{align}
where $cos(\cdot)$ is a cosine similarity. The $\nabla_{z_{l,c}} \mathcal{L}_{kd}$ and $\nabla_{z_{l,c}} \mathcal{L}_{cls}$ are the gradients of KD and classification losses for each channel $c$ for the features ${z_{l,c}}$ of the $l$-th layer, respectively. The binary mask $M$ is activated when directions of the two gradients for the new class data are equal. If the mask is activated, feature distillation is applied. Using the new class data, pattern completion is achieved through KD when the gradients of KD and classification losses have same directions for each channel of the features. On the other hands, applying KD can inhibit the classification performance when the two gradients have different directions~\cite{cos}. Therefore, in this case, we rule out the KD to achieve the pattern separation. Nonetheless, not regularizing at all for a particular channel may cause problems with the stability of the model. Additionally, in the process of pattern separation, from the perspective of pursuing differentiation from existing knowledge, maintaining the existing pattern can lead to complete differentiation. To achieve this, we apply a mask that is opposite to the binary mask to the old classes data while applying the binary mask to the new data.

Finally, the objective for our GFD is defined as follows:
\begin{align}
\mathcal{L}_{gfd} = M\cdot\mathcal{L}_{fd}(x_{new}) + \left(1-M\right)\cdot\mathcal{L}_{fd}(x_{old}),
\end{align}
where,
\begin{align}
\mathcal{L}_{fd} = \sum_{l=1}^L \sum_{c=1}^C \left|\left| z^t_{l,c} - z^{t-1}_{l,c}\right|\right|_F^2,
\end{align}
in which $x_{new}$ and $x_{old}$ are new classes data and old classes data, respectively. $L$ is the total number of layers and $C$ is the total number of channels in each layer. $\left|\left| \cdot \right|\right|_F$ is the frobenius norm and $\mathcal{L}_{fd}$ is a channel-wise feature distillation loss, and we normalize each feature for learning stability.

\subsection{Confidence-aware Weight Interpolation}
Under the stability-plasticity dilemma~\cite{mermillod2013stability,verbeke2019learning} in the CIL, simultaneously improving stability and plasticity is a challenging problem. However, continual learning considering the balance between the stability and plasticity is essential to improve overall performance. Regarding stability-plasticity, previous methods~\cite{PODNet, AFC} have had difficulties in ensuring stability enough to expect satisfactory performance improvement despite performing feature distillation that directly transfer the representation. To overcome these limitations, we interpolate the weights of the old model into the new model, thereby preventing loss changes for the old data and ensuring stability for the previous task. In addition, we pursue the plasticity through exploratory learning after the stability of the model is guaranteed enough.

Consequently, we introduce a confidence-aware weight interpolation to improve both stability and plasticity. As the model learns on the new data, we interpolate the weights of the new model through the weights of the old model to prevent the model from getting out of the low error area due to a large difference between parameters with the old model. We update the parameters of the new model $\theta^t$ through the parameters of the old model $\theta^{t-1}$, which is expressed as:
\begin{align}
\theta^t \leftarrow \beta\theta^t + (1-\beta)\theta^{t-1},
\end{align}
where $\beta \in [0,1]$ is an interpolation parameter. However, weight interpolation can adversely affect plasticity from a continuous perspective. To solve this problem, we use confidence in the old data to enable exploratory learning without weight interpolation if the model has sufficient knowledge about the old data. We adaptively adjust the interpolation parameter $\beta$, to apply weight interpolation when the confidence of the new model and the old model for the old data is above the threshold, and not to apply when the confidence is below the threshold. Therefore, interpolation parameter $\beta$ is defined as follows:
\begin{align}
\beta = \begin{cases}
\alpha,\; if\; conf(x_{old};\theta^{t-1}) - conf(x_{old};\theta^{t}) \ge \delta \\
1,\; \mathrm{otherwise}
\end{cases},
\end{align}
in which confidence denotes $conf(x;\theta) = \mathbb{E} \left[p_y(x;\theta)\right]$ and $\alpha \in [0,1]$ is the hyperparmeter, where $p_y(x;\theta)$ is the predicted probability for the ground truth $y$, and $\delta$ is the threshold. Since the prediction distribution also changes as the class increases, we set $\delta = \lambda_{th} \delta^t$, where $\lambda_{th}$ is the hyperparameter, and adaptive factor is $\delta^t = \left(n^{t}\right)^2 / n^{1:t}$ and $n^{t}$ is the number of classes in task $t$.

\begin{figure}
\begin{center}
\includegraphics[width=7cm]{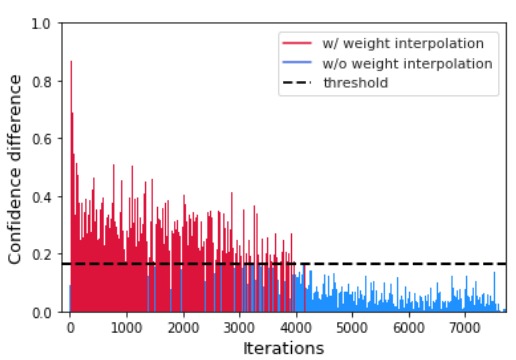}
\vspace{-0.4cm}
\end{center}
   \caption{Difference in confidence based on iterations. Our method pursues stability through weight interpolation in the unstable initial learning process when a new class is added to the model. In the subsequent learning process, which has stable performance on previously learned data, it pursues exploratory learning on a new class by eliminating regularization by weight interpolation.}
\label{fig:Figure3}
\vspace{-0.2cm}
\end{figure}

Figure~\ref{fig:Figure3} shows the difference in confidence between the old and new model for the previous 50 classes data in the process of learning an additional 10 classes from the model trained for 50 classes on CIFAR-100 as a result of applying CWI. In the initial learning process, while learning new class data, the prediction performance for old classes is destroyed, but the difference in confidence for old classes is reduced through feature distillation and weight interpolation. In addition, in the later learning process, when the difference in confidence between the two models is lower than the threshold, it is confirmed that a certain level of confidence is guaranteed by exploratory learning on new class data without weight interpolation.

\begin{algorithm}[t]
  \caption{SRIL: Selective Regularization Algorithm}
  \begin{algorithmic}[0]
  \State \textbf{Input:} training set $D^t$, exemplar set $\mathcal{E}^t$, old model's parameters $\theta^{t-1}$, new model's parameters $\theta^t$, interpolation parameter $\alpha$, distill factor $\lambda^t_{gfd}$, learning rate $\gamma$
  \State \textbf{Output:} $\theta^t$
  \State $\theta^t \leftarrow \theta^{t-1}$ // initialize new model
  \For{$e$ in $\{ 1, \cdots , E \}$}
    \State sample a mini-batch $(x,y) \sim D^t \cup \mathcal{E}^t$
      \State compute confidence from old classes $(x_{old},y_{old})$
      \If {$conf(x_{old};\theta^{t-1}) - conf(x_{old};\theta^{t}) \ge \delta $}
          \State $\theta^t \leftarrow \alpha\theta^t + (1-\alpha)\theta^{t-1}$
      \EndIf
      \State zero initialization binary mask $M$
      \State compute gradient from new classes $(x_{new},y_{new})$
      \For{$l$ in $\{ 1, \cdots , L \}$}
        \If { cos($\nabla_{z_{l,c}} \mathcal{L}_{kd} , \nabla_{z_{l,c}} \mathcal{L}_{cls})\ge 0$}
          \State $M_{l,c} = 1$
        \EndIf
      \EndFor
      \State $\mathcal{L}_{gfd} = M\cdot\mathcal{L}_{fd}(x_{new}) + \left(1-M\right)\cdot\mathcal{L}_{fd}(x_{old})$
      \State $\mathcal{L}_{total} = \mathcal{L}_{cls} + \lambda^t_{gfd} \mathcal{L}_{gfd}$
      \State  compute gradient $\nabla \mathcal{L}_{total}$
      \State  update parameters $\theta^t \leftarrow \theta^t -\gamma \nabla \mathcal{L}_{total}$
  \EndFor
  \State update exemplar from $D^t \cup \mathcal{E}^t$
  \end{algorithmic}
\end{algorithm}

\subsection{Overall Framework}
Our SRIL consists of Gradient-based Feature Distillation and Confidence-aware Weight Interpolation. GFD is a KD method for considering different expressive features of the old class and new class. CWI balances between stability and plasticity. Our overall loss is follow as:
\begin{align}
\mathcal{L}_{total} = \mathcal{L}_{lsc} + \lambda^t_{gfd}\mathcal{L}_{gfd},
\end{align}
in which distill factor is $\lambda^t_{gfd} = \lambda_{gfd} \cdot \lambda^t$, where $\lambda_{gfd}$ is a hyperparameter, and  adaptation factor is $\lambda^t = \sqrt{n^{1:t} / n^t}$. $\lambda^t$ is a well-used adaptation factor in CIL~\cite{LUCIR, PODNet,GeoDL, AFC}. We also use local similarity classifiers (LSC) following the previous works~\cite{PODNet,AFC}. LSC allows K proxies for each class. The logit for each class $\mathsf{c}$ is obtained as the average of K proxies.
\begin{align}
s_{\mathsf{c},k} = \frac{ exp \left\langle \phi_{\mathsf{c},k},z \right\rangle}{\sum_i exp \left\langle \phi_{\mathsf{c},i},z \right\rangle} \quad \hat{\mathbf{y}}_\mathsf{c} = \sum_{k} s_{\mathsf{c},k} 	\left\langle \phi_{\mathsf{c},k},z \right\rangle,
\end{align}
where $\phi$ is a classifier weight, $s_{\mathsf{c},k}$ and $\hat{\mathbf{y}}_\mathsf{c}$ denote the probability by each proxy and the probability for each class. We use LSC loss~\cite{PODNet} which is based on NCA loss~\cite{proxynca} as follows:
\begin{align}
\mathcal{L}_{lsc} = \left[ -\log \frac{{ exp (\eta (\hat{\mathbf{y}}_y - \varepsilon}))}{\sum_{i \neq y}exp (\eta\hat{\mathbf{y}}_i)} \right]_+,
\end{align}
where $\eta$ and $\varepsilon$ are learnable parameter and small margin, respectively, and hinge $[\cdot]_+$ denotes $max(0,\cdot)$.

\section{Experiment}
To verify the experimental validity, we compare performance with the state-of-the-art replay-based methods that share the same problem as ours and use KD. We perform experiments using the small scale CIFAR-100 dataset and the large scale ImageNet dataset, both of which are generally used in CIL. Exemplar stores 20 fixed samples for each class with herding selection~\cite{ICaRL}. We report both results of CNN predictions and nearest-meaning-of-exemplars~\cite{ICaRL} classification, denoted as CNN and NME, respectively.

\begin{table*}[t]
\begin{center}
\caption{Comparison of average accuracy (\%) between state-of-the-art methods using exempler and our proposed SRIL on CIFAR-100. The best accuracy is indicated in bold.}
\vspace{2mm}
\label{tab:main-result1}
\begin{tabular}{l|cccc}
\hline
\multicolumn{1}{c}{} &\multicolumn{4}{c}{CIFAR-100}\\
 & 50 tasks & 25 tasks & 10 tasks & 5 tasks \\ 
New classes per task & 1 & 2 & 5 & 10 \\ \hline
iCaRL~\cite{ICaRL}           & $44.20\pm0.98$  & $50.60\pm1.06$  & $53.78\pm1.16$  & $58.08\pm0.59$   \\
BiC~\cite{BiC}             & $47.09\pm1.48$  & $48.96\pm1.03$  & $53.21\pm1.01$  & $56.86\pm0.46$   \\
LUCIR (NME)~\cite{LUCIR}           & $48.57\pm0.37$  & $56.82\pm0.19$  & $60.83\pm0.70$  & $63.63\pm0.87$   \\
LUCIR (CNN)~\cite{LUCIR}           & $49.30\pm0.32$  & $57.57\pm0.23$  & $61.22\pm0.69$  & $64.01\pm0.91$   \\
PODNet (NME)~\cite{PODNet}    & $61.40\pm0.68$  & $62.71\pm1.26$  & $64.03\pm1.30$  & $64.48\pm1.32$   \\
PODNet (CNN)~\cite{PODNet}    & $57.98\pm0.46$  & $60.72\pm1.36$  & $63.19\pm1.16$  & $64.83\pm0.98$   \\
GeoDL~\cite{GeoDL}           & $52.28\pm3.91$  & $60.21\pm0.46$  & $63.61\pm0.81$  & $65.34\pm1.05$   \\
DDE~\cite{DDE}           & -  & -  & $64.12\pm1.40$  & $65.42\pm0.72$   \\
AANet~\cite{AANet}           & -  & $62.31\pm1.02$  & $64.31\pm0.90$  & $66.31\pm0.87$   \\
AFC (NME)~\cite{AFC}       & $62.58\pm1.02$  & $64.06\pm0.73$  & $64.29\pm0.92$  & $65.82\pm0.88$   \\
AFC (CNN)~\cite{AFC}       & $62.18\pm0.57$  & $63.89\pm0.93$  & $64.98\pm0.87$  & $66.49\pm0.81$   \\
CSCCT~\cite{CSCCT}           & $58.80\pm1.92$  & $61.10\pm1.12$  & $63.72\pm1.06$  & -   \\ \hline
SRIL (NME)      & $\mathbf{63.84\pm0.98}$  & $\mathbf{64.87\pm0.91}$  & $\mathbf{66.25\pm1.16}$  & $\mathbf{67.13\pm1.03}$   \\
SRIL (CNN)      & $63.64\pm1.02$  & $64.36\pm1.16$  & $65.11\pm0.82$  & $66.21\pm0.89$   \\ \hline
\end{tabular}
\end{center}
\vspace{-0.5cm}
\end{table*}

\subsection{Experimental Setup}

\textbf{Dataset and protocol.} We evaluate our method on CIFAR-100~\cite{CIFAR}, ImageNet-Subset~\cite{ImageNet,LUCIR,PODNet} and ImageNet-Full~\cite{ImageNet}. CIFAR-100 consists of 60,000 images of $32 \times 32$ pixels with 100 classes. It contains 500 training data and 100 test data for each class. ImageNet is a large-scale classification dataset, including 1.28 million images and 50k test set, and consists of 1,000 classes. ImageNet-Full uses all classes of ImageNet, and ImageNet-Subset refers to a dataset that randomly extracts 100 classes out of 1,000 classes. We use the same random seed and class order as the previous methods~\cite{PODNet,AFC} for fair comparison. After learning a part of the entire class for CIL setting, we gradually learn a certain number of classes. Feature-based distillation methods transfer representation directly, so representation of previously learned models is important. Therefore, it shows better performance than response-based distillation in small task incremental learning settings, where it learns about half of the total class in advance and then gradually learns a small number of classes~\cite{LUCIR, PODNet, AFC}.

\textbf{Implementation details.} For reproducibility, we conducted an experiment based on the PODNet~\cite{PODNet} code. All experiments are conducted on TITAN-Xp GPU and two GPUs are used for ImageNet-Full. We follow the experimental setup of PODNet~\cite{PODNet}. In the CIFAR-100 experiment, we use the ResNet-32~\cite{ResNet} architecture. We trained the model for 160 epochs with SGD with momentum of 0.9 and used a batch size of 128 and a weight decay of 0.0005. We use a cosine annealing learning rate scheduler with an initial learning rate of 0.1. The hyperparameters are $\alpha = 0.995, \lambda_{th}=0.1$, and $\lambda_{gfd} = 2$. For ImageNet-Subset and ImageNet-Full, we use the ResNet-18~\cite{ResNet} architecture. We trained the model for 90 epochs with SGD with momentum of 0.9 and used a cosine annealing learning rate scheduler with an initial learning rate of 0.1 and 0.05, respectively. We use a batch size of 64 and a weight decay of 0.0001. $\alpha$ and $\lambda_{th}$ use 0.999 and 0.1. $\lambda_{gfd}$ uses 5 and 7 for ImageNet-Subset and ImageNet-Full, respectively.

\subsection{Main Results}
\textbf{CIFAR-100.} We compare the average accuracy with the state-of-the-art methods that use the exemplar and KD~\cite{ICaRL, BiC, LUCIR, PODNet, GeoDL, DDE,AANet, AFC, CSCCT} in Table~\ref{tab:main-result1}. For a method based on other framework, GeoDL~\cite{GeoDL}, DDE~\cite{DDE}, AANet~\cite{AANet} and CSCCT~\cite{CSCCT} are a performance result based on PODNet (CNN)~\cite{PODNet}. We learn 50 classes in $0$-th task and increase the classes of fixed numbers for each task. As a result of the experiment using the NME classifier, the performance improvement of 0.81\text{--}1.96 percent points in all experimental environments. Experiments using a CNN classifier show that 5 tasks had lower performance of 0.28 percent points than closest state-of-the-art, but 10 tasks, 25 tasks and 50 tasks show performance improvements of 0.13\text{--}1.46 percent points compared to the state-of-the-art method.

\textbf{ImageNet-Subset/Full.} Unlike CIFAR-100, ImageNet uses only the CNN classifier for all experimental results based on the empirical fact that the performance of the knn-based NME classifier is poor due to the large number of classes. For the same reason, previous studies also only report CNN results for ImageNet experiments~\cite{PODNet,AFC}. Also, for methods that are additionally applied to other frameworks, results based on frameworks recorded with best performance. Table~\ref{tab:main-result2} shows the results of our ImageNet experiment. In the ImageNet-Subset, our method achieves state-of-the-art in all experiments, with performance 0.78\text{--}1.73 percent points higher than the nearest best performance. Experiments on ImageNet-Full achieved state-of-the-art by a difference of 0.07 percent point in experiments on 5 tasks, and 0.15 percent point lower than the state-of-the-art in experiments on 10 tasks.

\begin{table*}[t]
\begin{center}
\caption{Comparison of average accuracy (\%) between state-of-the-art methods and SRIL on ImageNet-Subset and ImageNet-Full. The best accuracy is represented in bold, and the second-best accuracy is underlined.}
\vspace{1mm}
\label{tab:main-result2}
\begin{tabular}{l|cccc|cc}
\hline
\multicolumn{1}{c}{} &\multicolumn{4}{c}{ImageNet-Subset}&\multicolumn{2}{c}{ImageNet-Full}\\
 & 50 tasks & 25 tasks & 10 tasks & 5 tasks  & 10 tasks & 5 tasks \\ 
Class per tasks & 1 & 2 & 5 & 10 & 50 & 100 \\ \hline
iCaRL~\cite{ICaRL}           & $54.97$  & $54.56$  & $60.90$  & $65.56$ & $46.72$ & $51.36$   \\
BiC~\cite{BiC}             & $46.49$  & $59.65$  & $65.14$  & $68.97$ & $44.31$ & $45.72$   \\
LUCIR~\cite{LUCIR}           & $55.44$  & $60.81$  & $65.83$  & $67.07$ & $59.92$ & $64.34$   \\
Mnemonics~\cite{Mnemonics}           & $-$  & $69.74$  & $71.37$  & $72.58$ & $63.01$ & $64.54$   \\
PODNet~\cite{PODNet}    & $62.48$  & $68.31$  & $74.33$  & $75.54$ & $64.13$ & $66.95$   \\
DDE~\cite{DDE}           & $-$  & $-$  & $75.41$  & $76.71$ & $64.71$ & $66.42$ \\
GeoDL~\cite{GeoDL}           & $-$  & $71.72$  & $73.55$  & $73.87$ & $64.46$ & $65.23$   \\
AANet~\cite{AANet}           & $-$  & $71.78$  & $75.58$  & \underline{76.96} & $64.85$ & $67.73$   \\
CSCCT~\cite{CSCCT}           & $-$  & $68.91$  & $74.35$  & $76.41$ & $-$ & $-$ \\
AFC~\cite{AFC}       & \underline{72.08}  & \underline{73.34}  & \underline{75.75}  & 76.84 & $\mathbf{67.02}$ & \underline{68.90}   \\ \hline
SRIL (Ours)      & $\mathbf{72.86}$  & $\mathbf{75.33}$  & $\mathbf{77.28}$  & $\mathbf{78.57}$ & \underline{66.87} & $\mathbf{68.97}$   \\ \hline
\end{tabular}
\end{center}
\vspace{-0.5cm}
\end{table*}

\subsection{Ablation studies}

\begin{figure}
\begin{center}
\includegraphics[width=8.5cm]{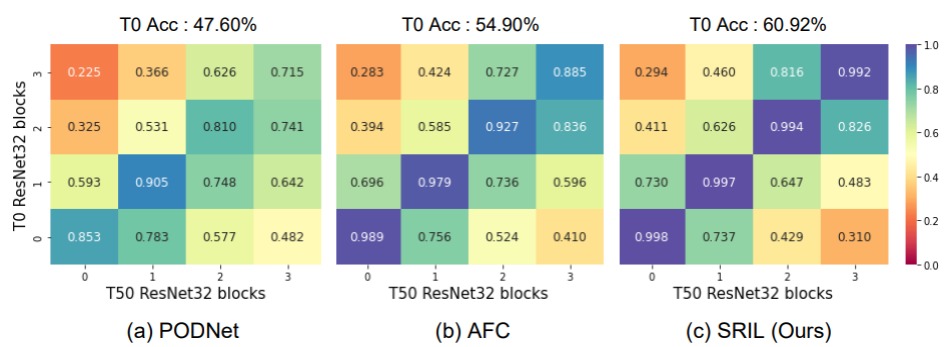}
\vspace{-0.5cm}
\end{center}
   \caption{Comparison of CKA similarity with other feature distillation methods.}
\label{fig:Figure4}
\vspace{-0.2cm}
\end{figure}

\textbf{Representation similarity analysis.} 
Figure~\ref{fig:Figure4} shows centered kernel alignment (CKA)~\cite{CKA} similarity for the intermediate feature between $0$-th task model and $50$-th task model trained on CIFAR-100. Both axis refers to the output of the middle layer, which applies feature distillation of the $0$-th task model and $50$-th task model. The diagonal of the heatmap is CKA similarity for the output of the same layer. Therefore, the closer the diagonal component is to 1, the higher representation similarity of the two models. As a method of using feature distillation, we compare PODNet~\cite{PODNet} and AFC~\cite{AFC} with our proposed method, SRIL. As a result, SRIL has higher CKA similarity than the existing methods. In particular, for the last intermediate feature closest to the embedding, the CKA similarities of PODNet, AFC and SRIL are 0.715, 0.885 and 0.992, respectively. T0 Acc at the top of Figure~\ref{fig:Figure4} denotes the accuracy of the $0$-th task data of the $50$-th task model. T0 Acc are $47.60\%$, $54.90\%$, and $60.92\%$, respectively, and SRIL achieves the best accuracy. Through this, we confirm that SRIL maintains performance for old classes and has high stability by preserving the representation.

\begin{figure}
\begin{center}
\includegraphics[width=7cm]{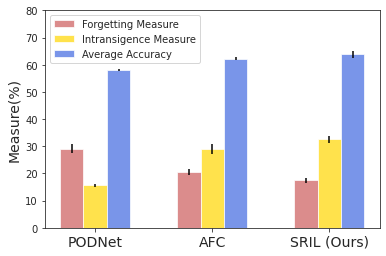}
\vspace{-0.4cm}
\end{center}
   \caption{Comparison of stability and plasticity with other replay-based methods.}
\label{fig:Figure5}
\vspace{-0.3cm}
\end{figure}

\textbf{Stability-Plasticity analysis.} We analyze Forgetting Measure ($FM$)~\cite{RWALK} and Intransigence Measure ($IM$)~\cite{RWALK} to determine how the stability and plasticity of SRIL differ from other feature distillation methods in Figure~\ref{fig:Figure5}. $FM$ is a measure of stability and is defined as the average difference between the maximum accuracy of the previous model and the accuracy of the current model for all tasks. A lower $FM$ value indicates higher stability. $IM$ is a measure of plasticity and is defined as an average difference in accuracy for new data between models learned without regularization  and models with regularization of all tasks. A lower $IM$ value indicates higher plasticity. As a result, PODNet~\cite{PODNet} has strengths in plasticity, and AFC~\cite{AFC} and SRIL have strengths in stability. Due to the stability-plasticity dilemma, no method is good for both $FM$ and $IM$, and our method strikes a good balance between stability and plasticity and achieves the best average accuracy. We explain the results of the main experiment through the results of stability-plasticity analysis. Our method has high stability by using two selective regularization methods, feature distillation and weight interpolation, and improves overall performance by maintaining performance on the old task. As a result, there was a large performance gain in small task setting, where the class increased by a small number. On ImageNet-Full, we achieved a similar level of performance to the existing method.

\textbf{Effect of each component.} We perform an ablation study to examine the effectiveness of each component of our methods in Table~\ref{tab:ablation1}. The separated feature distillation (SFD) is a general feature distillation applied to each of old and new classes, and is the same as the form without a mask in GFD. As a result of the experiment, in the case of GFD, there was an overall performance improvement when using the mask, and when using CWI, it shows a large performance improvement for the experiment learning 50 tasks, which are small tasks. In addition, it can be confirmed that the highest performance is obtained when the two methods are used together. Especially, in the case of the second column, stability and plasticity are not balanced due to lack of stability why only using feature distillation. However, by balancing stability and plasticity through CWI, we achieve a significant performance gain of 4.43 percent point.

\begin{table}
\begin{center}
\caption{Effect of each component with 50 tasks and 5 tasks on CIFAR-100.}
\vspace{-2mm}
\label{tab:ablation1}
\vspace{0.1cm}
\resizebox{\columnwidth}{!}{%
\begin{tabular}{ccc|cccc}
\hline
\multicolumn{3}{c|}{} & \multicolumn{2}{c}{50 tasks} & \multicolumn{2}{c}{5 tasks} \\ \hline
SFD & GFD & CWI   & NME & CNN &NME& CNN \\ \hline
    &   &   & $35.80\pm2.53$& $40.02\pm1.80$& $46.75\pm0.87$& $45.76\pm0.88$ \\
\checkmark &   &   & $62.48\pm0.50$  & $58.81\pm0.85$  & $66.07\pm0.77$  & $65.29\pm0.69$ \\
 & \checkmark &   & $62.89\pm1.03$  & $59.21\pm0.67$  & $66.41\pm1.14$  & $65.34\pm0.75$ \\   
\checkmark &   & \checkmark & $63.30\pm0.58$  & $63.27\pm1.11$  & $66.12\pm0.83$  & $65.73\pm1.40$ \\
 & \checkmark  & \checkmark & $\mathbf{63.84\pm0.98}$  & $\mathbf{63.64\pm1.02}$  & $\mathbf{67.13\pm1.03}$  & $\mathbf{66.21\pm0.89}$ \\ \hline
\end{tabular}%
}
\end{center}
\vspace{-6mm}
\end{table}

\begin{figure*}[t]
\begin{center}
\includegraphics[width=17cm]{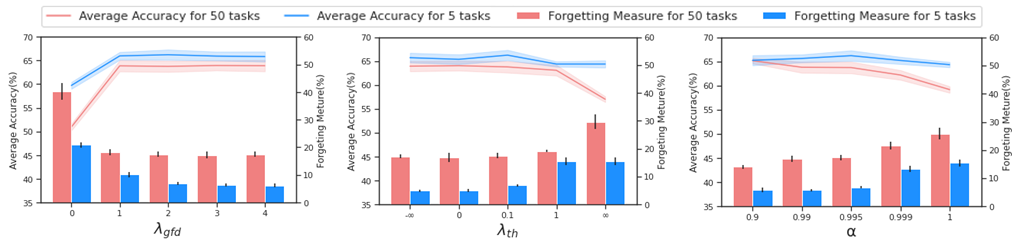}
\vspace{-0.5cm}
\end{center}
   \caption{Sensitivity analysis for each hyperparameter with 5 tasks and 50 tasks on CIFAR-100. The lineplot and barplot represent average accuracy and forgetting measure, respectively.}
\label{fig:Figure6}
\vspace{-0.4cm}
\end{figure*}

\textbf{Hyperparameter sensitivity analysis.} In Figure~\ref{fig:Figure6}, we investigate our hyperparameters $\lambda_{gfd}$ for the gradient-based feature distillation, and $\lambda_{th}$ and $\alpha$ for the confidence-aware weight interpolation on CIFAR-100. As $\lambda_{gfd}$ increases, the forgetting measure decreases, and the average accuracy remains similar. $FM$ increases as $\lambda_{th}$ and $\alpha$ increase. This means that stability decreases and plasticity increases as the hyperparameters of the two values increase. When the two values are infinity and 1, weight interpolation is not used at all, and at this time, the performance deteriorates the most. Our method can control the balance between plasticity and stability through these three parameters. Based on these results, we determine the final hyperparameters.

\begin{table}
\begin{center}
\caption{Effect of memory budget on CIFAR-100 with 50 tasks.$\ast$ Reproduced by the official code.}
\vspace{-2mm}
\label{tab:ablation2}
\resizebox{\columnwidth}{!}{%
\begin{tabular}{l|cccc}
\hline
Memory per class & 5 & 10 & 20 & 50 \\ \hline
iCaRL~\cite{ICaRL}                & $16.44$  & $28.57$  & $44.20$  & $48.29$  \\
BiC~\cite{BiC}                    & $20.84$  & $21.97$  & $47.09$  & $55.01$   \\
LUCIR (NME)~\cite{LUCIR}           & $21.81$  & $41.92$  & $48.57$  & $56.09$  \\
LUCIR (CNN)~\cite{LUCIR}           & $22.17$  & $42.70$  & $49.30$  & $57.02$  \\
PODNet (NME)~\cite{PODNet}         & $48.37$  & $57.20$  & $61.40$  & $62.27$  \\
PODNet (CNN)~\cite{PODNet}         & $35.59$  & $48.57$  & $57.98$  & $63.69$  \\
AFC$^\ast$ (NME)~\cite{AFC}        & $44.33$  & $57.27$  & $62.33$  & $63.83$ \\
AFC (CNN)~\cite{AFC}               & $44.66$  & $55.87$  & $62.18$  & $65.07$ \\ \hline 
SRIL (NME)                         & $\mathbf{54.09\pm1.45}$  & $\mathbf{61.99\pm0.66}$  & $\mathbf{63.84\pm0.98}$  & $65.31\pm1.30$ \\ 
SRIL (CNN)                         & $50.23\pm1.05$  & $60.47\pm1.31$  & $63.64\pm1.02$  & $\mathbf{65.31\pm1.01}$ \\ \hline
\end{tabular}%
}
\end{center}
\vspace{-2mm}
\end{table}

\begin{figure}
\begin{center}
\includegraphics[width=8cm]{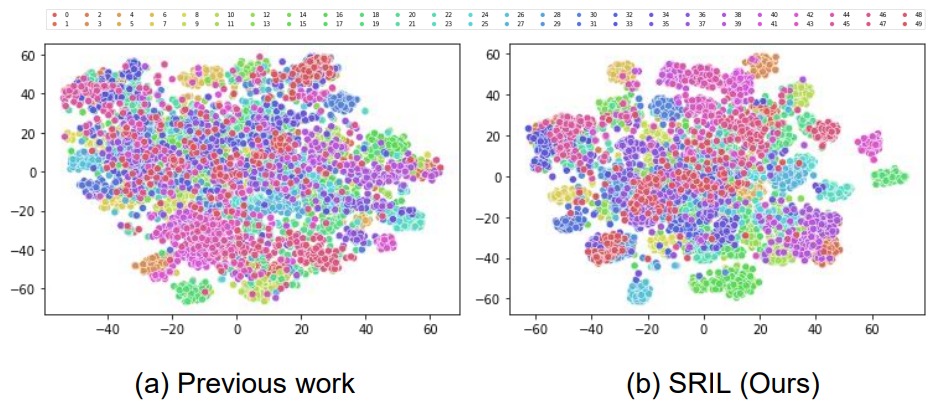}
\vspace{-0.4cm}
\end{center}
   \caption{t-SNE visualization of embedded features of the $0$-th task data for the model trained up to the $50$-th task on CIFAR-100.}
\label{fig:TSNE}
\vspace{-0.4cm}
\end{figure}

\textbf{Effect of memory budget.} We conduct an experiment on exemplar that stores 5, 10, 20, and 50 samples for each class to check the dependence by memory budget in Table~\ref{tab:ablation2}. Although our method adopts an asymmetric learning strategy for current task data and exemplar, our method achieves state-of-the-art for all memory sizes and shows robust performance even for small exemplar sizes. For the most challenging experiment with a memory size of 5 per class, we achieve significant performance improvements of 5.72 and 5.57 percent points for NME and CNN, respectively. 

\textbf{Feature visualization.} We visualize the embedded features of 0-th task data for model trained up to the $50$-th task on CIFAR-100 using t-SNE~\cite{TSNE} in Figure~\ref{fig:TSNE}. Compared to previous work~\cite{PODNet}, our method achieves to construct more cohesive clusters for each class. These results show that our method works successfully on knn based NME classifier and small exemplar size.

\textbf{Effect of initial task size.} To investigate the effect on the initial representation, we experiment by reducing the initial task size, the number of classes for the $0$-th task, from 50 to 20 by 10 in Table~\ref{tab:ablation3}. For all experiments, our method achieves state-of-the-art. Confirming good performance even in the case of lack of diversity in the initial representation demonstrates that our method is robust to the initial representation of the model.
\begin{table}
\begin{center}
\caption{Effect of varying initial task sizes on CIFAR-100 with 1 class per tasks.}
\vspace{-2mm}
\label{tab:ablation3}
\resizebox{\columnwidth}{!}{%
\begin{tabular}{l|cccc}
\hline
 & 80 tasks & 70 tasks & 60 tasks & 50 tasks \\
Initial task size & 20 & 30 & 40 & 50 \\ \hline
iCaRL~\cite{ICaRL}                & $41.28$  & $43.38$  & $44.35$  & $44.20$  \\
BiC~\cite{BiC}                    & $40.95$  & $42.27$  & $45.18$  & $47.09$   \\
LUCIR (NME)~\cite{LUCIR}           & $40.81$  & $46.80$  & $46.71$  & $48.57$  \\
LUCIR (CNN)~\cite{LUCIR}           & $41.69$  & $47.85$  & $47.51$  & $49.30$  \\
PODNet (NME)~\cite{PODNet}         & $49.03$  & $55.30$  & $57.89$  & $61.40$  \\
PODNet (CNN)~\cite{PODNet}         & $47.68$  & $52.88$  & $55.42$  & $57.98$  \\
AFC (NME)~\cite{AFC}               & $51.31$  & $57.05$  & $60.06$  & $62.58$ \\
AFC (CNN)~\cite{AFC}               & $52.90$  & $57.61$  & $60.27$  & $62.18$ \\ \hline 
SRIL (NME)                         & $\mathbf{53.21\pm1.42}$  & $58.30\pm1.49$  & $61.21\pm1.38$  & $\mathbf{63.84\pm0.98}$ \\ 
SRIL (CNN)                         & $53.18\pm1.11$  & $\mathbf{58.93\pm1.84}$  & $\mathbf{61.29\pm1.64}$  & $63.64\pm1.02$ \\ \hline
\end{tabular}%
}
\end{center}
\vspace{-0.5cm}
\end{table}

\section{Conclusion}
In this paper, we proposed a selective regularization method to accept new knowledge while preserving  previous knowledge. We introduced an asymmetric feature distillation approach using the gradients of classification and knowledge distillation losses to decide whether to perform pattern completion and separation. Furthermore, we proposed a confidence-aware weight interpolation method to improve the balance between stability and plasticity in class-incremental learning. We achieved competitive performance to the state-of-the-art methods, and extensive experiments demonstrate the effectiveness of our method in a variety of scenarios.

{\small
\bibliographystyle{ieee_fullname}
\bibliography{egbib}

\begin{thebibliography}{10}\itemsep=-1pt

\bibitem{MAS}
Rahaf Aljundi, Francesca Babiloni, Mohamed Elhoseiny, Marcus Rohrbach, and
  Tinne Tuytelaars.
\newblock Memory aware synapses: Learning what (not) to forget.
\newblock In {\em Proceedings of the European conference on computer vision
  (ECCV)}, pages 139--154, 2018.

\bibitem{CSCCT}
Arjun Ashok, KJ Joseph, and Vineeth~N Balasubramanian.
\newblock Class-incremental learning with cross-space clustering and controlled
  transfer.
\newblock In {\em Computer Vision--ECCV 2022: 17th European Conference, Tel
  Aviv, Israel, October 23--27, 2022, Proceedings, Part XXVII}, pages 105--122.
  Springer, 2022.

\bibitem{ERACE}
Lucas Caccia, Rahaf Aljundi, Nader Asadi, Tinne Tuytelaars, Joelle Pineau, and
  Eugene Belilovsky.
\newblock New insights on reducing abrupt representation change in online
  continual learning.
\newblock In {\em International Conference on Learning Representations}, 2022.

\bibitem{EEIL}
Francisco~M Castro, Manuel~J Marin-Jimenez, Nicolas Guil, Cordelia Schmid, and
  Karteek Alahari.
\newblock End-to-end incremental learning.
\newblock In {\em Proceedings of the European Conference on Computer Vision
  (ECCV)}, pages 233--248, 2018.

\bibitem{RWALK}
Arslan Chaudhry, Puneet~K Dokania, Thalaiyasingam Ajanthan, and Philip~HS Torr.
\newblock Riemannian walk for incremental learning: Understanding forgetting
  and intransigence.
\newblock In {\em ECCV}, 2018.

\bibitem{ImageNet}
Jia Deng, Wei Dong, Richard Socher, Li-Jia Li, Kai Li, and Li Fei-Fei.
\newblock Imagenet: A large-scale hierarchical image database.
\newblock In {\em 2009 IEEE conference on computer vision and pattern
  recognition}, pages 248--255. Ieee, 2009.

\bibitem{ERDIL}
Songlin Dong, Xiaopeng Hong, Xiaoyu Tao, Xinyuan Chang, Xing Wei, and Yihong
  Gong.
\newblock Few-shot class-incremental learning via relation knowledge
  distillation.
\newblock In {\em Proceedings of the AAAI Conference on Artificial
  Intelligence}, pages 1255--1263, 2021.

\bibitem{PODNet}
Arthur Douillard, Matthieu Cord, Charles Ollion, Thomas Robert, and Eduardo
  Valle.
\newblock Podnet: Pooled outputs distillation for small-tasks incremental
  learning.
\newblock In {\em Proceedings of the IEEE European Conference on Computer
  Vision (ECCV)}, 2020.

\bibitem{cos}
Yunshu Du, Wojciech~M Czarnecki, Siddhant~M Jayakumar, Mehrdad Farajtabar,
  Razvan Pascanu, and Balaji Lakshminarayanan.
\newblock Adapting auxiliary losses using gradient similarity.
\newblock {\em arXiv preprint arXiv:1812.02224}, 2018.

\bibitem{eeckt2022weight}
Steven~Vander Eeckt et~al.
\newblock Weight averaging: A simple yet effective method to overcome
  catastrophic forgetting in automatic speech recognition.
\newblock {\em ICASSP}, 2023.

\bibitem{Goodfellow13}
I.~J. Goodfellow, M. Mirza, D. Xiao, A. Courville, and Y. Bengio.
\newblock An empirical investigation of catastrophic forgetting in
  gradient-based neural networks.
\newblock {\em ArXiv e-prints}, dec 2013.

\bibitem{OCM}
Yiduo Guo, Bing Liu, and Dongyan Zhao.
\newblock Online continual learning through mutual information maximization.
\newblock In {\em International Conference on Machine Learning}, pages
  8109--8126. PMLR, 2022.

\bibitem{ResNet}
Kaiming He, Xiangyu Zhang, Shaoqing Ren, and Jian Sun.
\newblock Deep residual learning for image recognition.
\newblock In {\em Proceedings of the IEEE conference on computer vision and
  pattern recognition}, pages 770--778, 2016.

\bibitem{KD}
Geoffrey Hinton, Oriol Vinyals, and Jeff Dean.
\newblock Distilling the knowledge in a neural network.
\newblock {\em arXiv preprint arXiv:1503.02531}, 2015.

\bibitem{LUCIR}
Saihui Hou, Xinyu Pan, Chen~Change Loy, Zilei Wang, and Dahua Lin.
\newblock Learning a unified classifier incrementally via rebalancing.
\newblock In {\em The IEEE Conference on Computer Vision and Pattern
  Recognition (CVPR)}, June 2019.

\bibitem{DDE}
Xinting Hu, Kaihua Tang, Chunyan Miao, Xian-Sheng Hua, and Hanwang Zhang.
\newblock Distilling causal effect of data in class-incremental learning.
\newblock In {\em Proceedings of the IEEE/CVF conference on Computer Vision and
  Pattern Recognition}, pages 3957--3966, 2021.

\bibitem{CPG}
Ching-Yi Hung, Cheng-Hao Tu, Cheng-En Wu, Chien-Hung Chen, Yi-Ming Chan, and
  Chu-Song Chen.
\newblock Compacting, picking and growing for unforgetting continual learning.
\newblock In {\em Advances in Neural Information Processing Systems}, pages
  13647--13657, 2019.

\bibitem{SWA}
P Izmailov, AG Wilson, D Podoprikhin, D Vetrov, and T Garipov.
\newblock Averaging weights leads to wider optima and better generalization.
\newblock In {\em 34th Conference on Uncertainty in Artificial Intelligence
  2018, UAI 2018}, pages 876--885, 2018.

\bibitem{AFC}
Minsoo Kang, Jaeyoo Park, and Bohyung Han.
\newblock Class-incremental learning by knowledge distillation with adaptive
  feature consolidation.
\newblock In {\em Proceedings of the IEEE/CVF conference on computer vision and
  pattern recognition}, pages 16071--16080, 2022.

\bibitem{EWC}
James Kirkpatrick, Razvan Pascanu, Neil Rabinowitz, Joel Veness, Guillaume
  Desjardins, Andrei~A Rusu, Kieran Milan, John Quan, Tiago Ramalho, Agnieszka
  Grabska-Barwinska, et~al.
\newblock Overcoming catastrophic forgetting in neural networks.
\newblock {\em Proc. of the national academy of sciences}, 2017.

\bibitem{CKA}
Simon Kornblith, Mohammad Norouzi, Honglak Lee, and Geoffrey Hinton.
\newblock Similarity of neural network representations revisited.
\newblock In {\em International Conference on Machine Learning}, pages
  3519--3529. PMLR, 2019.

\bibitem{CIFAR}
Alex Krizhevsky, Geoffrey Hinton, et~al.
\newblock Learning multiple layers of features from tiny images.
\newblock 2009.

\bibitem{LwF}
Zhizhong Li and Derek Hoiem.
\newblock Learning without forgetting.
\newblock {\em IEEE Transactions on Pattern Analysis and Machine Intelligence},
  2017.

\bibitem{AANet}
Yaoyao Liu, Bernt Schiele, and Qianru Sun.
\newblock Adaptive aggregation networks for class-incremental learning.
\newblock In {\em Proceedings of the IEEE/CVF conference on Computer Vision and
  Pattern Recognition}, pages 2544--2553, 2021.

\bibitem{RMM}
Yaoyao Liu, Bernt Schiele, and Qianru Sun.
\newblock Rmm: Reinforced memory management for class-incremental learning.
\newblock {\em Advances in Neural Information Processing Systems},
  34:3478--3490, 2021.

\bibitem{Mnemonics}
Yaoyao Liu, Yuting Su, An-An Liu, Bernt Schiele, and Qianru Sun.
\newblock Mnemonics training: Multi-class incremental learning without
  forgetting.
\newblock In {\em Proceedings of the IEEE/CVF conference on Computer Vision and
  Pattern Recognition}, pages 12245--12254, 2020.

\bibitem{mccloskey1989catastrophic}
Michael McCloskey and Neal~J Cohen.
\newblock Catastrophic interference in connectionist networks: The sequential
  learning problem.
\newblock In {\em Psychology of learning and motivation}, volume~24, pages
  109--165. Elsevier, 1989.

\bibitem{mermillod2013stability}
Martial Mermillod, Aur{\'e}lia Bugaiska, and Patrick Bonin.
\newblock The stability-plasticity dilemma: Investigating the continuum from
  catastrophic forgetting to age-limited learning effects, 2013.

\bibitem{proxynca}
Yair Movshovitz-Attias, Alexander Toshev, Thomas~K Leung, Sergey Ioffe, and
  Saurabh Singh.
\newblock No fuss distance metric learning using proxies.
\newblock In {\em Proceedings of the IEEE international conference on computer
  vision}, pages 360--368, 2017.

\bibitem{patternseparation1}
Randall~C O'Reilly and James~L McClelland.
\newblock Hippocampal conjunctive encoding, storage, and recall: Avoiding a
  trade-off.
\newblock {\em Hippocampus}, 4(6):661--682, 1994.

\bibitem{ICaRL}
Sylvestre-Alvise Rebuffi, Alexander Kolesnikov, Georg Sperl, and Christoph~H
  Lampert.
\newblock icarl: Incremental classifier and representation learning.
\newblock In {\em Proceedings of the IEEE Conference on Computer Vision and
  Pattern Recognition}, pages 2001--2010, 2017.

\bibitem{PNN}
A.~A. Rusu, N.~C. Rabinowitz, G. Desjardins, H. Soyer, J. Kirkpatrick, K.
  Kavukcuoglu, R. Pascanu, and R. Hadsell.
\newblock Progressive neural networks.
\newblock {\em ArXiv e-prints}, jun 2016.

\bibitem{patternseparation2}
Adam Santoro.
\newblock Reassessing pattern separation in the dentate gyrus, 2013.

\bibitem{autonomous}
Khadija Shaheen, Muhammad~Abdullah Hanif, Osman Hasan, and Muhammad Shafique.
\newblock Continual learning for real-world autonomous systems: Algorithms,
  challenges and frameworks.
\newblock {\em Journal of Intelligent \& Robotic Systems}, 105(1):9, 2022.

\bibitem{GeoDL}
Christian Simon, Piotr Koniusz, and Mehrtash Harandi.
\newblock On learning the geodesic path for incremental learning.
\newblock In {\em Proceedings of the IEEE/CVF conference on Computer Vision and
  Pattern Recognition}, pages 1591--1600, 2021.

\bibitem{stojanovski2022momentum}
Zafir Stojanovski, Karsten Roth, and Zeynep Akata.
\newblock Momentum-based weight interpolation of strong zero-shot models for
  continual learning.
\newblock In {\em NeurIPS 2022 Workshop on Distribution Shifts: Connecting
  Methods and Applications}, 2022.

\bibitem{TPCIL}
Xiaoyu Tao, Xinyuan Chang, Xiaopeng Hong, Xing Wei, and Yihong Gong.
\newblock Topology-preserving class-incremental learning.
\newblock In {\em Computer Vision--ECCV 2020: 16th European Conference,
  Glasgow, UK, August 23--28, 2020, Proceedings, Part XIX 16}, pages 254--270.
  Springer, 2020.

\bibitem{TSNE}
Laurens Van~der Maaten and Geoffrey Hinton.
\newblock Visualizing data using t-sne.
\newblock {\em Journal of machine learning research}, 9(11), 2008.

\bibitem{verbeke2019learning}
Pieter Verbeke and Tom Verguts.
\newblock Learning to synchronize: How biological agents can couple neural task
  modules for dealing with the stability-plasticity dilemma.
\newblock {\em PLoS computational biology}, 15(8):e1006604, 2019.

\bibitem{modelsoups}
Mitchell Wortsman, Gabriel Ilharco, Samir~Ya Gadre, Rebecca Roelofs, Raphael
  Gontijo-Lopes, Ari~S Morcos, Hongseok Namkoong, Ali Farhadi, Yair Carmon,
  Simon Kornblith, et~al.
\newblock Model soups: averaging weights of multiple fine-tuned models improves
  accuracy without increasing inference time.
\newblock In {\em International Conference on Machine Learning}, pages
  23965--23998. PMLR, 2022.

\bibitem{WiSEFT}
Mitchell Wortsman, Gabriel Ilharco, Jong~Wook Kim, Mike Li, Simon Kornblith,
  Rebecca Roelofs, Raphael~Gontijo Lopes, Hannaneh Hajishirzi, Ali Farhadi,
  Hongseok Namkoong, et~al.
\newblock Robust fine-tuning of zero-shot models.
\newblock In {\em Proceedings of the IEEE/CVF Conference on Computer Vision and
  Pattern Recognition}, pages 7959--7971, 2022.

\bibitem{BiC}
Yue Wu, Yinpeng Chen, Lijuan Wang, Yuancheng Ye, Zicheng Liu, Yandong Guo, and
  Yun Fu.
\newblock Large scale incremental learning.
\newblock In {\em Proceedings of the IEEE/CVF Conference on Computer Vision and
  Pattern Recognition}, pages 374--382, 2019.

\bibitem{DEN}
J Yoon, E Yang, J Lee, and SJ Hwang.
\newblock Lifelong learning with dynamically expandable networks.
\newblock In {\em International Conference on Learning Representations, ICLR},
  2018.

\bibitem{SCKD}
Yichen Zhu and Yi Wang.
\newblock Student customized knowledge distillation: Bridging the gap between
  student and teacher.
\newblock In {\em Proceedings of the IEEE/CVF International Conference on
  Computer Vision}, pages 5057--5066, 2021.

\end{thebibliography}
}

\end{document}